\documentclass{INTERSPEECH2023}
\interspeechfinaltrue
\title{MD3: The Multi-Dialect Dataset of Dialogues}
\name{Jacob Eisenstein$^1$, Vinodkumar Prabhakaran$^1$, Clara Rivera$^1$ \\ Dorottya Demszky$^2$, Devyani Sharma$^3$}

\address{
  $^1$Google Research\\
  $^2$Stanford University \\
  $^3$Queen Mary University of London}
\email{\{jeisenstein, vinodkpg, rivera\}@google.com\\ddemszky@stanford.edu, d.sharma@qmul.ac.uk}

\newcommand{\say}[1]{\textit{#1}}
\usepackage{cleveref}
\usepackage{fancyvrb}
\usepackage{fvextra}
\fvset{breaklines=True,fontsize=\footnotesize,fontfamily=lmtt,commandchars=\\\{\}}
\usepackage{multirow}
\usepackage{comment}
\usepackage{url}

\usepackage{xcolor}

\definecolor{WowColor}{rgb}{.75,0,.75}
\definecolor{SubtleColor}{rgb}{0,0,.50}

\newcounter{margincounter}

\newcommand{\term}[1]{\textbf{#1}}

\begin{document}

\maketitle
 
\begin{abstract}
We introduce a new dataset of conversational speech representing English from India, Nigeria, and the United States. The Multi-Dialect Dataset of Dialogues (MD3) strikes a new balance between open-ended conversational speech and task-oriented dialogue by prompting participants to perform a series of short information-sharing tasks.
This facilitates quantitative cross-dialectal comparison, while avoiding the imposition of a restrictive task structure that might inhibit the expression of dialect features.
Preliminary analysis of the dataset reveals significant differences in syntax and in the use of discourse markers.
The dataset, which will be made publicly available with the publication of this paper, includes more than 20 hours of audio and more than 200,000 orthographically-transcribed tokens.\footnote{The MD3 dataset is publicly available at \href{https://www.kaggle.com/datasets/jacobeis99/md3en}{https://www.kaggle.com/datasets/jacobeis99/md3en}.}
\end{abstract}

\section{Introduction}
A key research challenge for spoken language processing is to build systems that meet users where they are: in their own languages and dialects. While there has been significant progress towards multi\emph{lingual} speech and text processing (e.g.,~\cite{pratap2020massively}), the development of multi\emph{dialectal} systems, datasets, and evaluations lags behind. Because billions of people speak dialects of global languages such as English, Arabic, and Spanish, multidialectal speech processing could dramatically increase access to language technology.

In this paper, we present MD3, the \textbf{M}ulti-\textbf{D}ialect \textbf{D}ataset of \textbf{D}ialogues. The current release of MD3 includes a total of 20 hours of audio from three varieties of global English: India, Nigeria, and the United States. Unlike most previous datasets of dialectal speech, which focused largely on scripted speech (e.g., \cite{arslan1996language}) or open-ended conversations (e.g., \cite{canavan1996callfriend}), MD3 is centered on information-sharing tasks with clearly-defined speaker intents. This makes it possible to study the dialect robustness of spoken-language processing systems not only in phonology but also in downstream language-processing tasks that are closely related to applications such as information retrieval and question answering. But unlike task-based dialogue scenarios, MD3 has no limitation on the vocabulary or grammatical structure, facilitating a more uninhibited style of interaction in which dialect features are more likely to appear~\cite{labov2006social}.

\begin{figure}
    \centering
    \includegraphics[width=.9\linewidth]{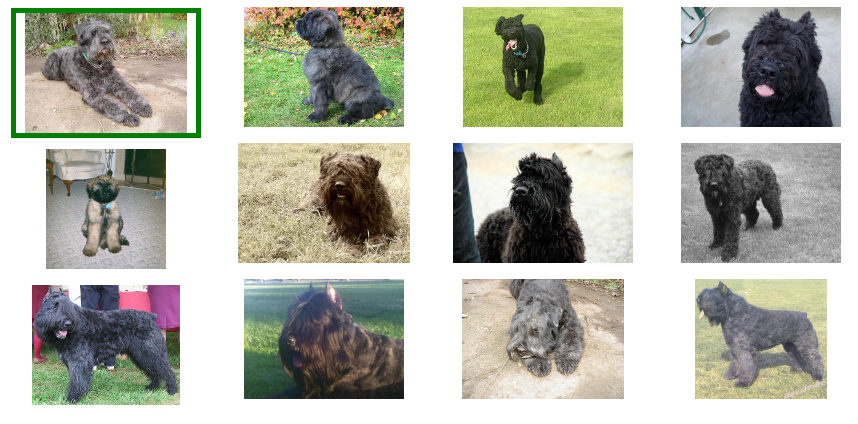}
\begin{Verbatim}
D: Here is a dog image. It is in grey colour. Ah uh um we can see one chain in his neck.
D: Background, we can see grass.
G: It is looking left side?
D: Ah it's looking left side but his face towards camera only.
G: Okay, okay. Done.
\end{Verbatim}
\caption{An example dialogue from two en-IN speakers. In the transcript, 'D' indicates the describer, who sees only the single image shown in the upper left; 'G' indicates the guesser, who sees all twelve images. The dialogue includes the en-IN dialect feature ``focus {\normalfont only}'' and the confirmation marker {\normalfont done} described in \cref{sec:dialect-features}.}
\label{fig:image-game-example}
\end{figure}

The MD3 conversations are organized around guessing games, in which one speaker (the ``describer") must communicate a piece of information to the other (the ``guesser"). There are two types of games: an image-guessing game (\Cref{fig:image-game-example}), in which the describer must describe an image well enough for the guesser to select it from a set of twelve similar images, and a word-guessing game (\Cref{fig:word-game-example}), in which the describer must communicate a word or phrase while avoiding a list of related words. This methodology elicits speech that is comparable in register and topic, without telling the participants what to say. The current release includes 3689 such games, orthographically transcribed into approximately 200,000 words (see \Cref{tab:statistics}). We also release metadata about the guessing games that prompted each dialogue. We hope that this dataset will serve as a benchmark for dialect-robust spoken language processing and as a resource for the study of global English.

\section{Related work}
Early work on ``accent classification'' focused on datasets of short scripted speech~\cite{arslan1996language}. Subsequent shared tasks on language identification included the classification of en-US and en-IN in spontaneous conversational telephone speech~\cite{le2005NIST}, using the CALLFRIEND corpora~\cite{canavan1996callfriend}. Multi-dialect datasets of conversational speech have been gathered in other languages, including Arabic~\cite{wray2015crowdsource}, Austrian~\cite{schuppler2014grass}, Swiss German~\cite{samardzic2016archimob}, and Spanish~\cite{zissman1996automatic}. Beyond dialect classification, researchers have explored the detection of specific dialect features~\cite{demszky-etal-2021-learning,masis-etal-2022-corpus} and the quantitative density of dialect features~\cite{craig2000assessment,koenecke2020racial,johnson2022automatic}. In the speech domain, such work has focused primarily on unconstrained narrative-based corpora such as the International Corpus of English~\cite{greenbaum1996international} and the Corpus of Regional African American Language~\cite{kendall2021corpus}. 

While unconstrained speech data has been a powerful resource for the study of dialect variation and the development of dialect-robust speech \emph{recognition}, it is less ideal for the development of dialect-robust \emph{speech processing systems}. This genre of speech often touches on local entities, such as place names, which are strong signals to the locale of the conversation while carrying little interesting dialectological information~\cite{demszky-etal-2021-learning}. Furthermore, unconstrained conversation is not directly applicable to the typical tasks of interest for speech processing systems, such as semantic parsing and question answering. The MD3 corpus addresses the first issue by restricting conversation to a predetermined set of topics governed by the information-sharing games that were used as conversational prompts. Regarding applicability to speech processing tasks, we have tried to strike a middle ground between unconstrained conversation and task-oriented dialogue systems (e.g.,~\cite{budzianowski-etal-2018-multiwoz}), which are likely to inhibit dialect features by imposing a rigid task model. The information-sharing tasks in MD3 are not derived from any specific application, but are related to challenging queries in search and image retrieval. Most importantly, accuracy can be measured at the level of individual prompts, making it possible for users of the dataset to extend fairness analyses from prior work on speech recognition (e.g.,~\cite{tatman2017effects,koenecke2020racial}) to downstream components of the speech processing pipeline.

\section{Elicitation}
\label{sec:elicitation}
The elicitation was performed in parallel in three locales: the United States (en-US), India (en-IN), and Nigeria (en-NG). In each locale, we constructed a pool of speakers from which we selected random pairs of individuals for a set of \term{matches}. No individual speaker participated in more than six matches and no pair participated in more than a single match.
Each match was divided into five \term{rounds}, in which one speaker was given the role of \term{describer}, and the other speaker was given the role of \term{guesser} (see \Cref{fig:image-game-example}). 
Within each round, the guesser and describer received a series of role-specific \term{prompts}, which required the describer to convey information to the guesser.
Each round was five minutes long, and the participants received randomly-selected prompts until the time expired.
The same pool of prompts was used in each locale. 

The elicitation procedure was subject to an internal review, ensuring that we obtained informed consent from the participants and protected their privacy.

\subsection{Speakers}
Speakers were recruited by two third-party vendors. In each locale, the vendor recruited an equal number of female and male participants. All participants were at least 18 years old. Geographically, we targeted three broad and diverse geographical regions. To ensure a level of linguistic coherence within each region, we imposed additional demographic criteria:
\begin{itemize}
    \item en-IN: native speaker of Telugu; high proficiency in English; recruited in Hyderabad.
    \item en-NG: native speaker of Yoruba; brought up and educated in English.
    \item en-US: native speakers of English; born in the Western United States (U.S. Census region 4, district 9). 
\end{itemize}
Demographic criteria were self-reported, so we cannot be completely certain that all speakers meet these criteria. Note that there is considerable linguistic heterogeneity within each group: for example, the en-US subcorpus includes speakers of African-American English. 

\begin{figure}
    \centering
    \hrule
    \begin{description}
    \item[Target] Science Fiction
    \item[Taboo words] Future, Imaginary, Advancements, Time, Travel
    \end{description}
    \hrule
    \begin{Verbatim}
D: Emm do you know when you are in secondary school, you you are when you get to like ss1 there are three classes that you can be. It's either you are in, do you know those three classes?
G: Commercial.
D: Sorry?
G: Commercial, Science, Art.
D: Ok. You know the middle one you mentioned,
G: which one?
D: the one the middle one you mentioned, 
G: science
D: yea hold it now, do you know when they say yea. do you know when you are watching a movie and the movie is not real what do you call that kind of movie?
G: fiction.
D: Yea so, join the two.
G: Science fiction.
\dots
\end{Verbatim}
    \caption{An example dialogue from two en-NG speakers, with 'D' indicating the describer and 'G' indicating the guesser. The describer uses a Nigerian cultural reference ({\normalfont ss1}) to introduce the term {\normalfont science}, and they later solve the full clue.}
    \label{fig:word-game-example}
\end{figure}

\subsection{Prompts}
Participants played two types of guessing games: an image-guessing game and a word-guessing game.
In the image-guessing game, the guesser is shown twelve similar images (see \Cref{fig:image-game-example}), one of which is also shown to the describer.
The participants must discuss what they see until the guesser can identify the describer's image;
if the guesser clicks on the correct image, the prompt is marked as a \term{win}, otherwise it is marked as a \term{loss}.
The word-guessing game is similar to the popular game ``Taboo'': the goal is for the guesser to identify a given term known to the describer, who may not use that term nor any of a set of five ``taboo terms.'' An example is shown in \Cref{fig:word-game-example}.
If the guesser states the target term, the prompt is marked as a win; if the describer accidentally uses the target term or one of the forbidden terms, it is marked as a loss. 
In the word game, we rely on self-reports for these outcomes.
Participants were not given any incentive to successfully solve the prompts and had the option to \term{skip} any prompt. Nonetheless, as shown in \Cref{tab:prompt-results}, the participants solved most prompts successfully --- bearing in mind that in the word-guessing game, results are based only on self-reports.

\begin{table*}[]
\eightpt
    \centering
    \begin{tabular}{lrrrrrrrr}
\toprule
{} &  speakers &  dialogues &  rounds &  prompts &  utterances &  hours &  tokens & WER \\
\midrule
\textbf{en-IN} &        27 &         46 &     134 &     1103 &       11856 &   6.37 &   62318 & 33.2\\
\textbf{en-NG} &        39 &         44 &     124 &      957 &       11482 &   6.48 &   67237 & 38.6\\
\textbf{en-US} &        38 &         37 &     152 &     1629 &       13235 &   8.94 &   86314 & 22.3\\
\bottomrule
\end{tabular}
    \caption{Statistics of the dataset by locale. For details on the distinction between dialogues, rounds, and prompts, see \Cref{sec:elicitation}. \term{Tokens} are counted by simple whitespace delimiting. For details on the word error rate calculation, see \Cref{sec:asr-description}.}
    \label{tab:statistics}
\end{table*}

\subsubsection{Image prompts}
The image prompts are drawn from three public datasets: FoodX-251~\cite{kaur2019foodx}, CalTech-UCSD Birds~\cite{wah2011caltech}, and Stanford Dogs~\cite{khosla2011novel}. These datasets were chosen because they offer fine-grained image classes which are not easy to distinguish with a single word or phrase. From the Stanford Dogs dataset we showed the guesser twelve images of the same breed of dog; from the CalTech-UCSD Birds dataset we showed images of the same species of bird; and from FoodX-251 we showed images of the same type of food (e.g., falafel). As shown in \Cref{fig:image-game-example}, this restriction forces the participants to describe several visual details of the image, including the spatial arrangement, background, and colors.

\subsubsection{Word prompts}
The prompts for the word guessing game require the target words and the corresponding forbidden words. Target words were drawn from a number of sources: (1)
two publicly-available Taboo datasets: Tabouid~\cite{bernard-2020-tabouid} and ``NS'',\footnote{\url{https://github.com/nehasinha/Taboo/blob/master/assets/cards.csv}} (2) a curated list of popular personalities, actors, singers, athletes, buildings, and statues based on Wikipedia's popular pages listing,\footnote{\url{https://en.wikipedia.org/wiki/Wikipedia:Popular_pages}} (3) a list of the most frequent English words from Education First,\footnote{\url{https://www.ef.com/wwen/english-resources/english-vocabulary/top-3000-words/}}
(4) a manually-curated word list originally designed for the acquisition of text-to-speech data. From all the prompts, we removed entities that were unlikely to be widely known in all locales, names of living politicians, as well as offensive and sexual terms. 
For NS prompts, we used the forbidden words that were provided in the dataset. 
For prompts from all other sources, we elicited forbidden words through crowdsourcing. For each target word, three raters were asked to provide five most common words that they would use to describe it. We aggregated the responses by selecting the words that were most-commonly chosen across raters, breaking ties manually.

\begin{table}[]
\eightpt
    \centering
\begin{tabular}{llrrr}
\toprule
locale & game type   &  win &  loss &  skip \\
\midrule
\multirow{2}{*}{en-IN} & image & 90.6 &   9.4 &   0.0 \\
      & word & 77.5 &   9.6 &  12.9 \\[6pt]
\multirow{2}{*}{en-NG} & image & 86.1 &  13.3 &   0.5 \\
      & word & 67.8 &   8.5 &  23.7 \\[6pt]
\multirow{2}{*}{en-US} & image & 91.3 &   8.7 &   0.0 \\
      & word & 83.7 &   4.1 &  12.0 \\
\bottomrule
\end{tabular}

    \caption{Results for prompts by locale and type. In the image game, a loss occurs when the guesser clicks on the wrong image; in the word game, a loss is when the describer accidentally uses the target term or one of the forbidden words.}
    \label{tab:prompt-results}
\end{table}

\subsection{Recording}
At the time the dataset was recorded, COVID-19 restrictions made it impossible for participants to meet in person.
Instead, they joined virtual meetings using Google Meet (they were asked to turn off their cameras), and simultaneously logged on to a crowdwork interface that presented the prompts.\footnote{Before their first match, all participants attended a group training, where they were provided written guidelines describing the guessing games rules with an overview of the technologies to be used during the recording process, and engaged in a supervised training round.}
Most participants worked from their homes.
Participants recorded their conversations using a proprietary web interface that securely stored and transmitted the audio.
No special audio equipment was provided; most participants used their laptop microphones and speakers, but some used headsets.
Many of the recordings contain background noise. 
We excluded audio files in which it was not possible to transcribe both participants. 
Audio is stored in 16-bit linear PCM encoding at 48 kHz in wav format. 

\subsection{Transcription}
We transcribed a subset of audio recordings in which both participants were clearly audible, relying on crosstalk for one of the two participants.
Orthographic transcription was performed by speech transcription professionals using audio files that were segmented by prompt.
In nearly every case, transcription was performed by an individual from the same geographic locale as the speakers, e.g., en-NG speakers were transcribed by workers in Nigeria.
Each trancription was reviewed by a second worker for accuracy.

\subsection{Locale differences}
While the form of the elicitation was identical across locales, there were significant differences in practice. 
Many of these differences were due to the fact that the speakers were communicating remotely from their own homes, using their own equipment.
In Nigeria, power outages caused a number of matches to be cancelled, and internet access was slow and unreliable. 
In the U.S., there was less background noise in the recordings, and speakers may have had access to higher-quality microphones.
For these reasons, although we recorded an identical number of dialogues in each locale, the final collection includes 30\% more audio time in en-US than the other two locales (see \Cref{tab:statistics}). 

A second area of difference relates to the prompts. Despite our efforts to identify prompts that would be solvable in all locales --- as well as in Spanish and Portuguese-speaking locales that will be included in a future dataset release --- some prompts were clearly more difficult for the non-US speakers. The issue was more significant in the word prompts, which were skipped nearly twice as often in the en-NG subcorpus as in en-IN and en-US (see \Cref{tab:prompt-results}). Differences were smaller in the image game: although the transcripts indicate that participants were sometimes unfamiliar with the types of food shown, this was rarely necessary to solve the game because in each prompt all of the candidate images were the same type of food. 

The US-based speakers completed prompts at a higher rate than the other two locales ($3.0$ prompts / minute in the U.S., $2.9$ in India, and $2.5$ in Nigeria). This could be attributed to both of the above factors (familiarity and technological resources), as well as other possible causes such as English proficiency and cultural differences. 

\section{Dataset}
Basic statistics of the dataset are shown in \Cref{tab:statistics}. Roughly the same number of dialogues were recorded in all three locales, but as noted above, a greater proportion of the en-US dialogues were of sufficiently high audio quality to transcribe.

\subsection{Speech recognition}
\label{sec:asr-description}
As a first test of the difficulty of the dataset for speech recognition, we applied the Whisper speech recognition system, using the \textsc{small.en} checkpoint~\cite{radford2022robust}. Word error rates (WER) are shown in the rightmost column of \Cref{tab:statistics}, ranging from 22.3 in en-US to 38.6 in en-NG. This indicates that the dataset is relatively difficult: in the Whisper paper, only two of fourteen datasets yield a higher error rate than the MD3 en-US subcorpus (CHiME6 and AMI-SDM1), and none has a higher error than the MD3 en-NG subcorpus. 
Potential causes for these differences include dialect sensitivity of the Whisper model as well as the different levels of audio quality in each locale.

\subsection{Dialect features}
\label{sec:dialect-features}
Goal-oriented interactive tasks help to divert attention from the recording situation and encourage naturalistic speech. As a result, the MD3 subcorpora exhibit many natural dialect features that can support a wide range of fine-grained analysis of language variation. For example, the speech samples in en-NG and en-IN instantiate naturalistic usage of classic phonological characteristics of Nigerian English~\cite{gut-2017-ng}, particularly those associated with Yoruba speakers, and Indian English~\cite{pingali-2009-in}, particularly those associated with Telugu speakers. Similarly, there is extensive variation in lexical, morphosyntactic, and dialogic features in the three subcorpora. We illustrate this systematic diversity with a snapshot of three syntactic and dialogic features in MD3, each with a distinct predicted distribution across the three dialects. The MD3 counts for these features are summarized in \Cref{tab:dialect-features}.

\begin{table}[]
    \centering
    \eightpt
    \begin{tabular}{p{2.2cm}lrrr}
\toprule
Feature & Meaning/form & en-IN & en-NG & en-US \\
\midrule
\say{Having} as  
& standard & 1 & 1 & 2\\
stative progressive & extended & 30 & 11 & 0 \\[16pt]
Focus \say{only}
& exclusive & 193 & 31 & 42\\
& focus & 23 & 0 & 0 \\[12pt]
Confirmation
& \say{okay} [all] & 1362 & 543 & 1734\\
 markers & \say{got it} [all] & 435 & 209  & 367 \\
& \say{okay got it} & 75 & 2  & 35\\
& \say{gotcha} & 0 & 0 & 13 \\
& \say{done} & 219 & 6 & 1\\
\bottomrule
\end{tabular}
    \caption{Dialect feature counts. For more details, see \Cref{sec:dialect-features}.}
    \label{tab:dialect-features}
\end{table}

The first feature is the use of progressive \say{-ing} with extended stative meanings, e.g. \say{it’s actually having its tongue slightly out} [en-NG]. Extended use of progressive \say{-ing} has been attested robustly for Indian English speakers~\cite{sharma-2009-in} and also for Yoruba English speakers at a lower rate~\cite{gutfuchs-2013-ng}, due to shared grammatical properties in the local languages (L1s) of the two regions. By contrast, it is not a feature of American English. This is precisely instantiated in the corpus: 30 of 31 uses of \say{having} in en-IN are extended stative uses. These are also attested in en-NG, but at a lower rate (11 instances). As expected, there are no instances in en-US.  

The second feature is the use of \emph{only} with non-contrastive, presentational focus meaning rather than exclusive meaning, e.g. \say{Is it a chocolate cake? Yes yes chocolate cake only} [en-IN]. Unlike extended progressives, the L1 source of focus \say{only} arises in Indian languages~\cite{lange-2007-in} but not in Yoruba, and so is only predicted for en-IN, not en-NG and en-US, which should pattern together. We find that 23 of 216 instances of \say{only} in en-IN are associated with the novel, non-contrastive presentational focus meaning. As predicted, neither en-NG (31 instances of \say{only}) nor en-US (42 instances of \say{only}) have any such usage. Here en-IN stands apart not only in terms of two distinct meanings, but also in the overall frequency of use of \say{only}, also due to the prevalence of pragmatic markers in Indian languages.

Finally, discourse markers that affirm shared knowledge or agreement~\cite{taguchi-2002-dms} are very prevalent in the dataset given the nature of the task, e.g. \say{gotcha} [en-US]. These confirmation markers show fine-grained patterns of overlap and difference across varieties. All three subcorpora show high levels of use of \say{okay} and \say{got it}, but each also shows distinctive behaviors: en-IN includes 219 uses of \say{done} as an affirmation marker, with only 1 and 6 uses respectively in en-US and en-NG. En-US has 13 instances of \say{gotcha}, a form absent in the two other dialects. And notably, en-IN shows a much higher overall use of confirmation markers than the other dialects, while en-NG uses half the overall amount used in en-US. 

These examples show that dialects are not monoliths; they are calibrated combinations of features that vary in frequency depending on factors such as region and speech task. Though not examined here, each region also exhibits inter-speaker variation. The orderly variation present in the closely parallel speech samples of the MD3 corpus represents a unique new resource for both dialect-robust spoken language processing and for the analysis of global English varieties.

\subsection{Limitations}
While the dataset demonstrates meaningful dialectal variation, researchers should be cautious when drawing generalizations about the speech patterns of the represented dialects or their speakers. First, the dataset includes speakers with one of three first languages: Telugu (India), Yoruba (Nigeria) and English (US). This is only a small subset of first languages spoken in these countries. Many of the dialect features are specific to the first language of the speakers, and hence the linguistic patterns may not generalize to speakers with other first languages. Second, the recruitment process relied on third-party vendors, which may have introduced selection biases. Third, in the word guessing game, it was difficult to select prompts that worked equally well across locales: given the dominance of western culture in Wikipedia, it likely that these prompts still over-represent Western entities to some degree. Differential familiarity with these entities could elicit marginally different conversational patterns, which in turn might influence downstream properties of the dataset. Depending on the use case, users of this data may wish to consider supplementary sources to increase diversity and representation.

\section{Conclusion}
This paper presents the MD3 dataset, which includes several thousand conversational information-sharing dialogues from English speakers in India, Nigeria, and United States. A first-pass investigation of the dataset reveals significant linguistic differences across the three locales. The MD3 dataset is distinguished by two key design decisions. First, the focus on information-sharing tasks makes it possible to define the \emph{intent} of each dialogue. In future work we plan to test the accuracy and dialect robustness of systems for recovering this intent from the text of the conversation. The second key design decision is the focus on nation-level dialects of global English. Much prior work on robustness in speech recognition has focused on what Wassink et al. call ``ethnicity-related dialects''~\cite{wassink2022uneven}, such as African-American English. Our view is that global English is relatively understudied from a robustness perspective, and we hope that this dataset draws attention to this pervasive form of language variation. That said, the MD3 elicitation methodology could be directly applied to other classes of dialects, and we view this too as an interesting possibility for future work.

\ifinterspeechfinal
\vspace{2ex}
\noindent\textbf{Acknowledgments.} 
Jon Clark played a key early role in the design and execution of the elicitation.
The dataset collection received engineering and administrative support from 
Landis Baker, 
David Elworthy,
Daphne Luong,
Mohd Majeed,
Sunny Mak,
Ravi Rajakumar,
Slav Petrov,
and
Austin Tarango.
We received valuable advice on audio and speech processing from Abhinav Garg and Kevin Wilson.
The research also benefitted from feedback from
Vera Axelrod,
Jan Botha,
Jason Baldridge,
Tim Dozat,
Jason Riesa, and
Jiao Sun.
Special thanks to the research participants whose conversations make up the dataset.
\fi

\bibliographystyle{IEEEtran}
\bibliography{mybib}

\end{document}